\documentclass[11pt,a4paper]{article}
\usepackage[hyperref]{emnlp2020}
\usepackage{times}
\usepackage{latexsym}

\usepackage{microtype}

\usepackage[compact]{titlesec}
\usepackage{makecell}
\usepackage{xspace}
\usepackage{amsmath,amssymb}
\usepackage{multirow}
\usepackage{booktabs}
\usepackage{graphicx}
\usepackage{wrapfig}
\usepackage{algorithm,algorithmic}
\usepackage{eqparbox}
\renewcommand{\algorithmiccomment}[1]{\hfill\eqparbox{COMMENT}{\# #1}}

\aclfinalcopy 

\newcommand{\selfsup}{SMLMT\xspace}
\newcommand{\selfsupmulti}{Hybrid-SMLMT\xspace}

\title{Self-Supervised Meta-Learning \\
for Few-Shot Natural Language Classification Tasks} 

\author{
Trapit Bansal$\;^\diamondsuit$\thanks{\;\;Correspondence: \texttt{tbansal@cs.umass.edu}} \and Rishikesh Jha$^\dagger$ \and Tsendsuren Munkhdalai$^\ddagger$ \and Andrew McCallum$^\diamondsuit$ \\
$^\diamondsuit$University of Massachusetts, Amherst \\
$^\dagger$Code for Science and Society \\
$^\ddagger$Microsoft Research, Montr\'eal, Canada
}

\date{}

\begin{document}
\maketitle
\begin{abstract}
Self-supervised pre-training of transformer models has revolutionized NLP applications. Such pre-training with language modeling objectives provides a useful initial point for parameters that generalize well to new tasks with fine-tuning. However, fine-tuning is still data inefficient --- when there are few labeled examples, accuracy can be low. Data efficiency can be improved by optimizing pre-training directly for future fine-tuning with few examples; this can be treated as a meta-learning problem. However, standard meta-learning techniques require many training tasks in order to generalize; unfortunately, finding a diverse set of such supervised tasks is usually difficult. This paper proposes a self-supervised approach to generate a large, rich, meta-learning task distribution from unlabeled text. This is achieved using a cloze-style objective, but creating separate multi-class classification tasks by gathering tokens-to-be blanked from among only a handful of vocabulary terms. This yields as many unique meta-training tasks as the number of subsets of vocabulary terms. We meta-train a transformer model on this distribution of tasks using a recent meta-learning framework. On 17 NLP tasks, we show that this meta-training leads to better few-shot generalization than language-model pre-training followed by finetuning. Furthermore, we show how the self-supervised tasks can be combined with supervised tasks for meta-learning, providing substantial accuracy gains over previous supervised meta-learning.
\end{abstract}

\section{Introduction}
Self-supervised learning has emerged as an important training paradigm for learning model parameters which are more generalizable and yield better representations for many down-stream tasks.
This typically involves learning through labels that come naturally with data, for example words in natural language.
Self-supervised tasks typically pose a supervised learning problem that can benefit from lots of naturally available data and enable pre-training of model parameters that act as a useful prior for supervised fine-tuning \cite{erhan2010does}.
Masked language modeling \cite{devlin2018bert}, and other related approaches \cite{peters2018deep,howard2018universal,radford2019language},
is an example of such a self-supervised task that is behind the success of transformer models like BERT.

While self-supervised pre-training is beneficial, it has been recently noted that it is not data-efficient and typically requires large amounts of fine-tuning data for good performance on a target task \cite{yogatama2018learning, bansal2019learning}.
This can be evaluated as a few-shot learning problem, where a model is given only few examples of a new task and is expected to perform well on that task.
This paper focuses on this problem of few-shot learning and develops models which demonstrate better few-shot generalization to new tasks.

Large scale pre-training suffers from a train-test mismatch as the model is not optimized to learn an initial point that yields good performance when fine-tuned with few examples. Moreover, fine-tuning of a pre-trained model typically introduces new random parameters, such as softmax layers, and important hyper-parameters such as learning rate, which are hard to estimate robustly from the few examples.
Thus, we propose to remove this train-test mismatch, and treat learning an initial point
and hyper-parameters jointly from unlabelled data, which allows data-efficient fine-tuning, as a meta-learning problem.

Meta-learning, or learning to learn \cite{thrun2012learning,schmidhuber1987evolutionary}, treats learning a parameterized algorithm, such as a neural net optimized with SGD, that generalizes to new tasks as a learning problem. This typically assumes access to a distribution over tasks in order to enable learning.
Creating tasks which enable meta-learning is one of the main challenges for meta-learning \cite{bengio1992optimization, santoro2016meta, vinyals2016matching}, and typical supervised meta-learning approaches create task distributions from a fixed task dataset with large number of labels by sub-sampling from the set of labels \cite{vinyals2016matching,ravi2017optimization}.
While this enables generalization to new labels, it limits generalization to unseen tasks due to over-fitting to the training task distribution \cite{Yin2020Meta-Learning}.
Moreover, large supervised datasets with a large label set are not always available for meta-learning, as is often the case in many NLP applications.

To overcome these challenges of supervised meta-learning, we propose a self-supervised approach and create the task-distribution from unlabelled sentences. 
Taking inspiration from the cloze task \cite{taylor1953cloze}, we create separate multi-class classification tasks by gathering tokens-to-be blanked from a subset of vocabulary words, 
allowing for as many unique meta-training tasks as the number of subsets of words in the language.
The proposed approach, which we call Subset Masked Language Modeling Tasks (SMLMT), 
enables training of meta-learning methods for NLP at a much larger scale than was previously feasible while also ameliorating the risk of over-fitting to the training task distribution.
This opens up new possibilities for applications of meta-learning in NLP, such as few-shot learning, continual learning, architecture search and more.

This work focuses on few-shot learning and makes the following contributions:
(1) we introduce a self-supervised approach to create tasks for meta-learning in NLP, Subset Masked Language Modeling Tasks (SMLMT), which enables application of meta-learning algorithms for goals like few-shot learning;
(2) utilizing SMLMT as the training task distribution, we train a state-of-the-art transformer architecture, BERT \cite{devlin2018bert}, using a recent optimization-based meta-learning method which was developed for diverse classification tasks \cite{bansal2019learning};
(3) we show that the self-supervised SMLMT can also be combined with supervised task data to enable better feature learning, 
while still allowing for better generalization by avoiding meta-overfitting to the supervised tasks through the use of SMLMT;
(4) we rigorously evaluate the proposed approach on few-shot generalization to unseen tasks as well as new domains of tasks seen during training and show that the proposed approach demonstrates better generalization than self-supervised pre-training or self-supervised pre-training followed by multi-task training;
(5) we also study the effect of number of parameters for few-shot learning and find that while bigger pre-trained or meta-trained models generalize better than smaller models, meta-learning leads to substantial gains even for the smaller models.

\section{Preliminaries}
In supervised meta-learning, we typically assume access to a task distribution $\mathcal{P}(\mathcal{T})$. Practically, this translates to a fixed set of training tasks $\{T_1, \ldots, T_M\}$, which are referred to as meta-training tasks.
For supervised classification, each task $T_i$ is an $N_i$-way classification task.
While many meta-learning algorithms assume a fixed $N$-way classification, we follow the more practical approach of \citet{bansal2019learning} and allow for a diverse set of classification tasks with potentially different number of classes.

The goal of a meta-learning algorithm is to utilize the meta-training tasks to learn a learning procedure
that generalizes to held-out tasks $T' \sim \mathcal{P}(T)$.
Model-agnostic meta-learning (MAML) \cite{Finn:2017:MMF:3305381.3305498} is an example of such a meta-learning algorithm.
MAML learns an initial point $\theta$ for a classifier $f_{\theta}:x\rightarrow \hat{y}$, that can be optimized via gradient descent on the supervised loss $\mathcal{L}_i$ defined for the task $T_i$, using its support set $\mathcal{D}^{tr} \sim T_i$:
\begin{equation}
    \theta'_i \leftarrow \theta - \alpha \nabla_{\theta} \mathcal{L}_i(\mathcal{D}^{tr}, \theta) \label{eq:inner}
\end{equation}
where $\alpha$ is the learning rate.
The optimized point $\theta'$ is then evaluated on another validation set for the task, $D^{val} \sim T_i$, using the loss function $\mathcal{L}_i$. This loss across meta-training tasks serves as the training error to optimize the initial point and parameters like learning-rate ($\Theta:=\{\theta, \alpha\}$):
\begin{equation}
    \Theta \leftarrow \Theta - \beta \; \nabla_{\Theta} \mathbb{E}_{T_i \sim \mathcal{P}(\mathcal{T})} \left[ L_i(\mathcal{D}^{val}, \theta'_i) \right]
    \label{eq:outer}
\end{equation}
where $\beta$ is the learning rate for the meta-training process.
Training proceeds in an episodic framework \cite{vinyals2016matching}, where in each episode a mini-batch of tasks are sampled along with their support and validation sets, and the model parameters are optimized using \eqref{eq:inner} and \eqref{eq:outer}, which are also referred to as inner and outer loop, respectively.

\begin{figure}[t!]
    \centering
    \includegraphics[width=\linewidth]{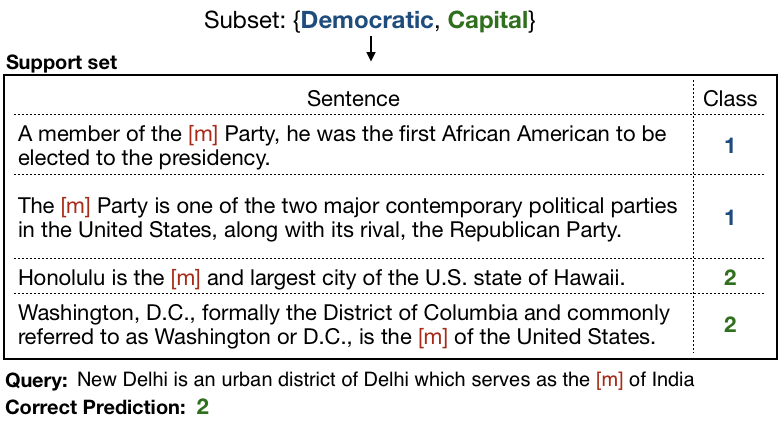}
    \caption{An example of a 2-way 2-shot task in SMLMT. The support set and one query is shown. Any $N$-way $k$-shot task can be constructed similarly.}
    \label{fig:task}
\end{figure}

\textbf{Meta-training Tasks:}
We summarize how supervised task data-sets are typically leveraged to create meta-training tasks \cite{vinyals2016matching}.
Assuming access to a supervised task with $L$ classes, an $N$-way $k$-shot task is created by first sampling $N$ classes, assuming $N << L$. Then for each of the $N$ sampled classes, $(k + q)$ examples of each class is randomly sampled from the dataset and assigned a unique label in $\{1, \ldots, N\}$.
The $k$ examples for each label serve as the support set, while the $q$ examples constitute the validation set described above.
Note, that each task consists of a small subset of classes and the class to label (1 to N) assignment is random . 
This is crucial to avoid learning the sample to label bindings in the parameters of the model, which will make the task-specific training (in \eqref{eq:inner}) irrelevant and the model will not generalize to new tasks.
An example of this approach is MiniImageNet \cite{ravi2017optimization}, which is a benchmark dataset for learning few-shot image classification.

\section{Self-supervised Tasks for Meta-learning}
The existing approach to using a supervised dataset to create tasks, as described above, is fraught with issues, specially for NLP applications.
First, note that large classification datasets with large label spaces are not readily available for all NLP tasks, for example sentiment classification which has only few discrete labels.
Second, limiting to a fixed supervised dataset to create tasks limits generalization ability and the meta-learned models might generalize to new labels for the task but fail to generalize to new novel tasks \cite{metz2018learning}.
Lastly, such an approach is also not feasible in all problems where we will like to apply meta-learning \cite{Yin2020Meta-Learning}. For example, consider meta-learning a natural language inference (NLI) model across multiple domains which can generalize to new domains. A powerful model can ignore the training data for each task and directly learn to predict the NLI tag for the examples in each training domain, which will lead to low training error but the model will not generalize to new domains.
We overcome these issues by utilizing unlabelled data to create meta-learning tasks.
See Fig.~\ref{fig:task} for an example of generated task.

\label{sec:smlm}
\textbf{Subset Masked Language Modeling Tasks (SMLMT):}
We are given a text corpus split into sentences $X_i$ and each sentence is a sequence of words from a vocabulary of size $V$.
Now, in Subset Masked Language Modeling Tasks, each task is defined from a \textit{subset} of vocabulary words. 
To create an $N$-way classification task, we randomly select $N$ unique vocabulary words: $\{v_1, \ldots, v_N\}$.
Then we consider all sentences containing these $N$ words, and for each word randomly sample $r=k+q$ sentences: $\textbf{x}_{v_i} = \{X_1, \ldots, X_{r} | v_i \in X_i\}$.
Now, we mask the corresponding chosen word from the sentences in each of these $N$ sets,
so $\textbf{x}'_{v_i} = \{\mbox{Mask}(X_1, v_i), \ldots, \mbox{Mask}(X_{r}, v_i)\}$ where $\mbox{Mask}(X, v)$ replaces all occurrences of $v$ in $X$ with the mask token $[m]$.
The set $\{\textbf{x}'_{v_1}, \ldots, \textbf{x}'_{v_N}\}$ is then a well-defined $N$-partition of $N\times r$ sentences, that serves as input examples for the $N$-way classification task.
We forget the original word corresponding to the masked tokens in these sets and assign labels in $\{1, \ldots, N\}$ to the $N$ sets.
This gives an instance of an SMLMT classification task: $T$ $=\{(x_{ij}, i) | i \in \{1, .. ,N\}, x_{ij} \in \textbf{x}'_{v_i} \}$.
This can be split into support and validation
for meta-training.

In an SMLMT instance, each input sentence consists of exactly one word that is masked throughout the sentence and its label corresponds to that word.
This requires a similar reasoning ability as cloze tasks \cite{taylor1953cloze}.
Moreover, crucially, the SMLMT task creation ensures that a model cannot memorize the input-label mapping as the target masked word is hidden and the label assignment is randomized, requiring the model to infer the labels from the support set.
Note that the SMLMT tasks are also closely related to masked language modeling (MLM) \cite{devlin2018bert}.
While MLM is a word-level classification task, SMLMT is a sentence-level classification task. 
Each unique subset of words from the vocabulary defines a unique task in SMLMT. This allows for as many unique tasks as the \textit{number of subsets of words in the vocabulary}, enabling large-scale 
meta-learning 
from unsupervised data.

\textbf{Hybrid SMLMT:}
Tasks from SMLMT can also be combined with supervised tasks to encourage better feature learning \cite{caruana1997multitask} and increase diversity in tasks for meta-learning.
We use a sampling ratio $\lambda \in (0,1)$ and in each episode select an SMLMT task with probability $\lambda$ or a supervised task with probability $(1-\lambda)$.
The use of SMLMT jointly with supervised tasks ameliorates meta-overfitting, as tasks in SMLMT cannot be solved without using the task support data.
$\lambda$ is a hyper-parameter. In our experiments, we found $\lambda=0.5$ to work well.


\section{Meta-learning Model}
\label{sec:model}
We now discuss the meta-learning model for learning new NLP tasks.

\textbf{Text encoder:} 
The input to the model is natural language sentences.
This is encoded using a transformer \cite{vaswani2017attention} text encoder. 
We follow the BERT \cite{devlin2018bert} model and use the same underlying neural architecture for the transformer as their base model.
Given an input sentence, the transformer model yields contextualized token representations for each token in the input after multiple layers of self-attention.
Following BERT, we add a special CLS token to the start of the input that is used as a sentence representation for classification tasks.
Given an input sentence $X$, let $f_{\pi}(X)$ be the CLS representation of the final layer of the transformer with parameters $\pi$.

\textbf{Meta-learning across diverse classes:}
Our motivation is to meta-learn an initial point that can generalize to novel NLP tasks, thus we consider methods that apply to diverse number of classes.
Note that many meta-learning models only apply to a fixed number of classes \cite{Finn:2017:MMF:3305381.3305498} and require training different models for different number of classes.
We follow the approach of \citet{bansal2019learning} that learns to generate softmax classification parameters conditioned on a task support set to enable training meta-learning models that can adapt to tasks with diverse classes.
This combines benefits of metric-based methods \cite{vinyals2016matching,snell2017prototypical} and optimization-based methods for meta-learning.
The key idea is to train a deep set encoder $g_{\psi}(\cdot)$, with parameters $\psi$, which takes as input the set of examples of a class $n$ and generates a $(d+1)$ dimensional embedding that serves as the linear weight and bias for class $n$ in the softmax classification layer.
Let $\{X_{1n}, \ldots, X_{kn}\}$ be the $k$ examples for class $n$ in the support set of a task $t$:
\begin{align}
    w_t^n, b_t^n &= g_{\psi}(\{f_{\pi}(X_{1n}), \ldots, f_{\pi}(X_{kn})\}) \label{eq:wandb} \\
    p(y|X) &= softmax\left\{ \textbf{W}_t\; h_{\phi}(f_{\pi}(X)) + \textbf{b}_t \right\} \label{eq:pred}
\end{align}
where $\textbf{W}_t = [w^1_t; \ldots; w^N_t] \in \mathcal{R}^{N\times d}$, $\textbf{b}_t = [b^1_t; \ldots; b^N_t] \in \mathcal{R}^d$ are the concatenation of the per-class vectors in \eqref{eq:wandb}, and $h_{\phi}$ is a MLP with parameters $\phi$ and output dimension $d$.

Using the above model to generate predictions, the parameters are meta-trained using the MAML algorithm \cite{Finn:2017:MMF:3305381.3305498}.
Concretely, set $\theta := \{\pi, \phi, \textbf{W}_t, \textbf{b}_t\}$ for the task-specific inner loop gradient updates in \eqref{eq:inner} and set $\Theta := \{\pi, \psi, \alpha\}$ for the outer-loop updates in \eqref{eq:outer}.
Note that we do multiple steps of gradient descent in the inner loop. \citet{bansal2019learning} performed extensive ablations over parameter-efficient versions of the model and found that adapting all parameters with learned per-layer learning rates performs best for new tasks.
We follow this approach.
Full training algorithm can be found in the Appendix.

\textbf{Fast adaptation:}
\citet{flennerhag2019meta} proposed an approach which mitigates slow adaption often observed in MAML
by learning to warp the task loss surface 
to enable rapid descent to the loss minima.
This is done by interleaving a neural network's layers with non-linear layers, called warp layers, which are not adapted for each task but are still optimized across tasks in the outer-loop updates in \eqref{eq:outer}.
Since introducing additional layers will make computation more expensive,
we use existing transformer layers as warp layers.
We designate the feed-forward layers in between self-attention layers of BERT,
which project from dimension 768 to 3072 to 768,
as warp-layers.
Note that these parameters also constitute a large fraction of total parameters ($\sim 51\%$). Thus in addition to the benefit from warping, not adapting these layers per task means significantly faster training and smaller number of per-task parameters during fine-tuning. The warp layers are still updated in the outer loop during meta-training.

\section{Related Work}
Language model pre-training has recently emerged as a prominent approach to learning general purpose representations \cite{howard2018universal,peters2018deep,devlin2018bert,radford2019language,yang2019xlnet,raffel2019exploring}. Refer to \citet{weng2019selfsup} for a review of self-supervised learning.
Pre-training is usually a two step process and fine-tuning introduces random parameters making it inefficient when target tasks have few examples \cite{bansal2019learning}.
Multi-task learning of pre-trained models has shown improved results on many tasks \cite{phang2018sentence,liu2019multi}. 
More recently, and parallel to this work, \citet{brown2020language} show that extremely large language models can act as few-shot learners. They propose a query-based approach where few-shot task data is used as context for the language model. In contrast, we employ a fine-tuning based meta-learning approach that enjoys nice properties like consistency which are important for good out-of-distribution generalization \cite{finn2018learning}.
Moreover, we show that self-supervised meta-learning can also improve few-shot performance for smaller models.

Meta-Learning methods can be categorized as: optimization-based \cite{Finn:2017:MMF:3305381.3305498,li2017meta, nichol2018reptile,rusu2018meta}, 
model-based \cite{santoro2016meta,ravi2017optimization,munkhdalai2017meta},
and metric-based \cite{vinyals2016matching,snell2017prototypical}.
Refer to \citet{finn2018learning} for an exhaustive review.
Unsupervised meta-learning has been explored in vision. \citet{hsu2018unsupervised} cluster images using pre-trained embeddings to create tasks for meta-learning.
\citet{metz2018learning} meta-learn a biologically-motivated update rule from unsupervised data
in a semi-supervised framework.
Compared to these, 
we directly utilize text data
to automatically create unsupervised tasks without relying on pre-trained embeddings or access to target tasks.

In NLP, meta-learning approaches have followed the recipe of using supervised task data and learning models for specific tasks.
Such approaches \cite{yu2018diverse,gu2018meta,guo2018multi,han2018fewrel,mi2019meta}
train to generalize to new labels of a specific task like relation classification and don't generalize to novel tasks.
\citet{bansal2019learning} proposed an approach that applies to diverse tasks to enable practical meta-learning models and evaluate on generalization to new tasks.
However, they rely on supervised task data from multiple tasks and suffer from meta-overfitting as we show in our empirical results.
\citet{holla2020learning} studied related approaches for the task of word-sense disambiguation.
To the best of our knowledge, the method proposed here is the first self-supervised approach to meta-learning in NLP.

\section{Experiments}
We evaluate the models on few-shot generalization to new tasks and new domains of train tasks.
Evaluation consist of a diverse set of NLP classification tasks from multiple domains: entity typing, sentiment classification, natural language inference and other text classification tasks.
Our results\footnote{Code and trained models: \url{https://github.com/iesl/metanlp}} show that self-supervised meta-learning using SMLMT improves performance over self-supervised pre-training.
Moreover, combining SMLMT with supervised tasks achieves the best generalization, improving over multi-task learning by up to 21\%.

\subsection{Implementation Details}

\textbf{SMLMT:}
We use the English Wikipedia dump, as of March 2019, to create SMLMT.
This is similar to the dataset for pre-training of BERT \cite{devlin2018bert}, which ensures that gains are not due to using more or diverse pre-training corpora \cite{liu2019roberta}.
The corpus is split into sentences and word-tokenized to create SMLMT.
We run task creation offline and create about 2 Million SMLMT for meta-training, including a combination of 2, 3 and 4-way tasks.
After task creation, the data is word-piece tokenized using the BERT-base cased model vocabulary for input to the models.

\textbf{Supervised Tasks:}
\citet{bansal2019learning} demonstrated that better feature learning from supervised tasks helps few-shot learning.
Thus, we also evaluate multi-task learning and multi-task meta-learning for few-shot generalization.
We also use GLUE tasks \cite{wang2018glue} and SNLI \cite{bowman2015snli} as the supervised tasks.
Supervised tasks can be combined with SMLMT for meta-training (see \ref{sec:smlm}).
Note that since these are only a few supervised tasks (8 in this case) with a small label space, it is easy for meta-learning models to overfit to the supervised tasks \cite{Yin2020Meta-Learning} limiting generalization as we show in experiments. 

\textbf{Models:}
We evaluate the following models:\\
(1) BERT: This is transformer model trained with self-supervised learning using MLM as the pre-training task on Wikipedia and BookCorpus. We use the cased base model \cite{devlin2018bert}.\\
(2) MT-BERT: This is a multi-task learning model trained on the supervised tasks. We follow \citet{bansal2019learning} in training this model. \\
(3) MT-BERT$_{\mbox{softmax}}$: This is the same model above where only the softmax layer is fine-tuned on downstream tasks. \\
(4) LEOPARD: This is the meta-learning model proposed in \citet{bansal2019learning} which is trained on only the supervised tasks.\\
(5) \selfsup: This is the meta-learning model (in \ref{sec:model}) which is trained on the self-supervised SMLMT. \\
(6) \selfsupmulti: This is the meta-learning model (in \ref{sec:model}) trained on a combination of SMLMT and supervised tasks.\\
Note that all models share the same transformer architecture making the contribution from each component discernible.
Moreover, \selfsup and \selfsupmulti models use similar meta-learning algorithm as LEOPARD, so any improvements are due to the self-supervised meta-training.
All model are initialized with pre-trained BERT for training.

\textbf{Evaluation Methodology:}
We evaluate on few-shot generalization to multiple NLP tasks using
the same set of tasks\footnote{Data: \url{https://github.com/iesl/leopard}} considered in \citet{bansal2019learning}. 
Each target task consists of $k$ examples per class, for $k \in \{4, 8, 16, 32\}$, and different tasks can have different number of classes.
Since few-shot performance is sensitive to the few examples used in fine-tuning, each model is fine-tuned on 10 such $k$-shot support sets for a task, for each $k$, and the average performance with standard deviation is reported.
Models are trained using their training procedures, without access to the target tasks, and are then fine-tuned for each of the $k$-shot task.
Results for MT-BERT and LEOPARD are taken from \citet{bansal2019learning}.

\paragraph{Hyper-parameters:}
We follow the approach of \citet{bansal2019learning} and use validation tasks for estimating hyper-parameters during fine-tuning for all baseline models.
Note the meta-learning approach learn the learning rates during training and only require the number of epochs of fine-tuning to be estimated from the validation tasks.
Detailed hyper-parameters are in Supplementary.

\subsection{Results}
\label{sec:results}
\subsubsection{Few-shot generalization to new tasks}
\label{sec:newtasks}
We first evaluate performance on novel tasks not seen during training.
The task datasets considered are: (1) entity typing: CoNLL-2003 \cite{sang2003conll}, MIT-Restaurant \cite{liu2013asgard}; (2) rating classification \cite{bansal2019learning}: 4 domains of classification tasks based on ratings from the Amazon Reviews dataset \cite{blitzer2007biographies}; (3) text classification: multiple social-media datasets from figure-eight\footnote{https://www.figure-eight.com/data-for-everyone/}.
\begin{table*}[htb!]
\centering \fontsize{8.0}{9.5}\selectfont \setlength{\tabcolsep}{0.5em}
\begin{tabular}{ccccc|cccc}
\Xhline{2\arrayrulewidth}
    Task     & $N$ & $k$ & BERT & \textbf{\selfsup} & $\mbox{MT-BERT}_{\mbox{softmax}}$ & MT-BERT & LEOPARD & \textbf{\selfsupmulti} \\[3pt] \Xhline{2\arrayrulewidth}
\multirow{4}{*}{CoNLL} & \multirow{4}{*}{4} & 4 & 50.44 \tiny{$\pm$ 08.57} & 46.81 \tiny{$\pm$ 4.77} & 52.28 \tiny{$\pm$ 4.06} & 55.63 \tiny{$\pm$ 4.99} & 54.16 \tiny{$\pm$ 6.32} & \textbf{57.60} \tiny{$\pm$ 7.11} \\
& & 8 & 50.06 \tiny{$\pm$ 11.30} & 61.72 \tiny{$\pm$ 3.11} &  65.34 \tiny{$\pm$ 7.12} & 58.32 \tiny{$\pm$ 3.77} & 67.38 \tiny{$\pm$ 4.33} & \textbf{70.20} \tiny{$\pm$ 3.00} \\
& & 16 & 74.47 \tiny{$\pm$ 03.10} & 75.82 \tiny{$\pm$ 4.04} & 71.67 \tiny{$\pm$ 3.03} & 71.29 \tiny{$\pm$ 3.30} & 76.37 \tiny{$\pm$ 3.08} & \textbf{80.61} \tiny{$\pm$ 2.77} \\
& & 32 & 83.27 \tiny{$\pm$ 02.14} & 84.01 \tiny{$\pm$ 1.73} & 73.09 \tiny{$\pm$ 2.42} & 79.94 \tiny{$\pm$ 2.45} & 83.61 \tiny{$\pm$ 2.40} & \textbf{85.51} \tiny{$\pm$ 1.73} \\
\midrule
\multirow{4}{*}{MITR} & \multirow{4}{*}{8} & 4 & 49.37 \tiny{$\pm$ 4.28} & 46.23 \tiny{$\pm$ 3,90} & 45.52 \tiny{$\pm$ 5.90} & 50.49 \tiny{$\pm$ 4.40} & 49.84 \tiny{$\pm$ 3.31} & \textbf{52.29} \tiny{$\pm$ 4.32} \\
& & 8 & 49.38 \tiny{$\pm$ 7.76} & 61.15 \tiny{$\pm$ 1.91} & 58.19 \tiny{$\pm$ 2.65} & 58.01 \tiny{$\pm$ 3.54} & 62.99 \tiny{$\pm$ 3.28} & \textbf{65.21} \tiny{$\pm$ 2.32} \\
& & 16 & 69.24 \tiny{$\pm$ 3.68} & 69.22 \tiny{$\pm$ 2.78} & 66.09 \tiny{$\pm$ 2.24} & 66.16 \tiny{$\pm$ 3.46} & 70.44 \tiny{$\pm$ 2.89} & \textbf{73.37} \tiny{$\pm$ 1.88} \\
& & 32 & 78.81 \tiny{$\pm$ 1.95} & 78.82 \tiny{$\pm$ 1.30} & 69.35 \tiny{$\pm$ 0.98} & 76.39 \tiny{$\pm$ 1.17} & 78.37 \tiny{$\pm$ 1.97} & \textbf{79.96} \tiny{$\pm$ 1.48} \\
\midrule
\multirow{4}{*}{Airline} & \multirow{4}{*}{3} & 4 & 42.76 \tiny{$\pm$ 13.50} & 42.83 \tiny{$\pm$ 6.12} & 43.73 \tiny{$\pm$ 7.86} & 46.29 \tiny{$\pm$ 12.26} & 54.95 \tiny{$\pm$ 11.81} & \textbf{56.46} \tiny{$\pm$ 10.67} \\
& & 8 & 38.00 \tiny{$\pm$ 17.06} & 51.48 \tiny{$\pm$ 7.35} & 52.39 \tiny{$\pm$ 3.97} & 49.81 \tiny{$\pm$ 10.86} & 61.44 \tiny{$\pm$ 03.90} & \textbf{63.05} \tiny{$\pm$ 8.25}\\
& & 16 & 58.01 \tiny{$\pm$ 08.23} & 58.42 \tiny{$\pm$ 3.44} & 58.79 \tiny{$\pm$ 2.97} & 57.25 \tiny{$\pm$ 09.90} & 62.15 \tiny{$\pm$ 05.56} & \textbf{69.33} \tiny{$\pm$ 2.24}\\
& & 32 & 63.70 \tiny{$\pm$ 4.40} & 65.33 \tiny{$\pm$ 3.83} & 61.06 \tiny{$\pm$ 3.89} & 62.49 \tiny{$\pm$ 4.48} & 67.44 \tiny{$\pm$ 01.22} & \textbf{71.21} \tiny{$\pm$ 3.28} \\
\midrule
\multirow{4}{*}{Disaster} & \multirow{4}{*}{2} & 4 & 55.73 \tiny{$\pm$ 10.29} & \textbf{62.26} \tiny{$\pm$ 9.16} & 52.87 \tiny{$\pm$ 6.16} & 50.61 \tiny{$\pm$ 8.33} & 51.45 \tiny{$\pm$ 4.25} & 55.26 \tiny{$\pm$ 8.32} \\
 & & 8 & 56.31 \tiny{$\pm$ 09.57} & \textbf{67.89} \tiny{$\pm$ 6.83} & 56.08 \tiny{$\pm$ 7.48} & 54.93 \tiny{$\pm$ 7.88} & 55.96 \tiny{$\pm$ 3.58} & 63.62 \tiny{$\pm$ 6.84} \\
 & & 16 & 64.52 \tiny{$\pm$ 08.93} & \textbf{72.86} \tiny{$\pm$ 1.70} & 65.83 \tiny{$\pm$ 4.19} & 60.70 \tiny{$\pm$ 6.05} & 61.32 \tiny{$\pm$ 2.83} & 70.56 \tiny{$\pm$ 2.23} \\
 & & 32 & 73.60 \tiny{$\pm$ 01.78} & \textbf{73.69} \tiny{$\pm$ 2.32} & 67.13 \tiny{$\pm$ 3.11} & 72.52 \tiny{$\pm$ 2.28} & 63.77 \tiny{$\pm$ 2.34} & 71.80 \tiny{$\pm$ 1.85} \\
\midrule
\multirow{4}{*}{Emotion} & \multirow{4}{*}{13} & 4 & 09.20 \tiny{$\pm$ 3.22} & 09.84 \tiny{$\pm$ 1.09} & 09.41 \tiny{$\pm$ 2.10} & 09.84 \tiny{$\pm$ 2.14} & 11.71 \tiny{$\pm$ 2.16} & \textbf{11.90} \tiny{$\pm$ 1.74} \\
 && 8 & 08.21 \tiny{$\pm$ 2.12} & 11.02 \tiny{$\pm$ 1.02} & 11.61 \tiny{$\pm$ 2.34} & 11.21 \tiny{$\pm$ 2.11} & 12.90 \tiny{$\pm$ 1.63} & \textbf{13.26} \tiny{$\pm$ 1.01} \\
 && 16 & 13.43 \tiny{$\pm$ 2.51} & 12.05 \tiny{$\pm$ 1.18} & 13.82 \tiny{$\pm$ 2.02} & 12.75 \tiny{$\pm$ 2.04} & 13.38 \tiny{$\pm$ 2.20} & \textbf{15.17} \tiny{$\pm$ 0.89} \\
 && 32 & 16.66 \tiny{$\pm$ 1.24} & 14.28 \tiny{$\pm$ 1.11} & 13.81 \tiny{$\pm$ 1.62} & \textbf{16.88} \tiny{$\pm$ 1.80} & 14.81 \tiny{$\pm$ 2.01} & 16.08 \tiny{$\pm$ 1.16} \\
\midrule
\multirow{4}{*}{Political Bias} & \multirow{4}{*}{2} & 4 & 54.57 \tiny{$\pm$ 5.02} & 57.72 \tiny{$\pm$ 5.72} & 54.32 \tiny{$\pm$ 3.90} & 54.66 \tiny{$\pm$ 3.74} & 60.49 \tiny{$\pm$ 6.66} & \textbf{61.17} \tiny{$\pm$ 4.91} \\
& & 8 & 56.15 \tiny{$\pm$ 3.75} & 63.02 \tiny{$\pm$ 4.62} & 57.36 \tiny{$\pm$ 4.32} & 54.79 \tiny{$\pm$ 4.19} & 61.74 \tiny{$\pm$ 6.73} & \textbf{64.10} \tiny{$\pm$ 4.03} \\
& & 16 & 60.96 \tiny{$\pm$ 4.25} & \textbf{66.35} \tiny{$\pm$ 2.84} & 59.24 \tiny{$\pm$ 4.25} & 60.30 \tiny{$\pm$ 3.26} & 65.08 \tiny{$\pm$ 2.14} & 66.11 \tiny{$\pm$ 2.04} \\
& & 32 & 65.04 \tiny{$\pm$ 2.32} & \textbf{67.73} \tiny{$\pm$ 2.27} & 62.68 \tiny{$\pm$ 3.21} & 64.99 \tiny{$\pm$ 3.05} & 64.67 \tiny{$\pm$ 3.41} & 67.30 \tiny{$\pm$ 1.53} \\
\midrule
\multirow{4}{*}{Political Audience} & \multirow{4}{*}{2} & 4 & 51.89 \tiny{$\pm$ 1.72} & \textbf{57.94} \tiny{$\pm$ 4.35} & 51.50 \tiny{$\pm$ 2.72} & 51.47 \tiny{$\pm$ 3.68} & 52.60 \tiny{$\pm$ 3.51} & 57.40 \tiny{$\pm$ 7.18} \\
 & & 8 & 52.80 \tiny{$\pm$ 2.72} & \textbf{62.82} \tiny{$\pm$ 4.50} & 53.53 \tiny{$\pm$ 2.26} & 54.34 \tiny{$\pm$ 2.88} & 54.31 \tiny{$\pm$ 3.95} & 60.01 \tiny{$\pm$ 4.54} \\
 & & 16 & 58.45 \tiny{$\pm$ 4.98} & \textbf{64.57} \tiny{$\pm$ 5.23} & 56.37 \tiny{$\pm$ 2.19} & 55.14 \tiny{$\pm$ 4.57} & 57.71 \tiny{$\pm$ 3.52} & 63.11 \tiny{$\pm$ 4.06} \\
 & & 32 & 55.31 \tiny{$\pm$ 1.46} & \textbf{67.68} \tiny{$\pm$ 3.12} & 53.09 \tiny{$\pm$ 1.33} & 55.69 \tiny{$\pm$ 1.88} & 52.50 \tiny{$\pm$ 1.53} & 65.50 \tiny{$\pm$ 3.78} \\
 \midrule
\multirow{4}{*}{Political Message} & \multirow{4}{*}{9} & 4 & 15.64 \tiny{$\pm$ 2.73} & 16.16 \tiny{$\pm$ 1.81} & 13.71 \tiny{$\pm$ 1.10} & 14.49 \tiny{$\pm$ 1.75} & 15.69 \tiny{$\pm$ 1.57} & \textbf{16.74} \tiny{$\pm$ 2.50} \\
 & & 8 & 13.38 \tiny{$\pm$ 1.74} & 19.24 \tiny{$\pm$ 2.32} & 14.33 \tiny{$\pm$ 1.32} & 15.24 \tiny{$\pm$ 2.81} & 18.02 \tiny{$\pm$ 2.32} & \textbf{20.33} \tiny{$\pm$ 1.22} \\
 & & 16 & 20.67 \tiny{$\pm$ 3.89} & 21.91 \tiny{$\pm$ 0.57} & 18.11 \tiny{$\pm$ 1.48} & 19.20 \tiny{$\pm$ 2.20} & 18.07 \tiny{$\pm$ 2.41} & \textbf{22.93} \tiny{$\pm$ 1.82} \\
 & & 32 & \textbf{24.60 \tiny{$\pm$ 1.81}} & 23.87 \tiny{$\pm$ 1.72} & 18.67 \tiny{$\pm$ 1.52} & 21.64 \tiny{$\pm$ 1.78} & 19.87 \tiny{$\pm$ 1.93} & 23.78 \tiny{$\pm$ 0.54} \\
\midrule
\multirow{4}{*}{Rating Electronics} & \multirow{4}{*}{3} & 4 & 39.27 \tiny{$\pm$ 10.15} & 37.69 \tiny{$\pm$ 4.82} & 39.89 \tiny{$\pm$ 5.83} & 41.20 \tiny{$\pm$ 10.69} & 51.71 \tiny{$\pm$ 7.20} & \textbf{53.74} \tiny{$\pm$ 10.17} \\
 & & 8 & 28.74 \tiny{$\pm$ 08.22} & 39.98 \tiny{$\pm$ 4.03} & 46.53 \tiny{$\pm$ 5.44} & 45.41 \tiny{$\pm$ 09.49} & 54.78 \tiny{$\pm$ 6.48} & \textbf{56.64} \tiny{$\pm$ 03.01} \\
 & & 16 & 45.48 \tiny{$\pm$ 06.13} & 45.85 \tiny{$\pm$ 4.72} & 48.71 \tiny{$\pm$ 6.16} & 47.29 \tiny{$\pm$ 10.55} & \textbf{58.69} \tiny{$\pm$ 2.41} & \textbf{58.67} \tiny{$\pm$ 03.73} \\
 & & 32 & 50.98 \tiny{$\pm$ 5.89} & 50.86 \tiny{$\pm$ 3.44} & 52.58 \tiny{$\pm$ 2.48} & 53.49 \tiny{$\pm$ 3.87} & 58.47 \tiny{$\pm$ 5.11} & \textbf{61.42} \tiny{$\pm$ 03.86} \\
\midrule
\multirow{4}{*}{Rating Kitchen} & \multirow{4}{*}{3} & 4 & 34.76 \tiny{$\pm$ 11.20} & 40.75 \tiny{$\pm$ 7.33} & 40.41 \tiny{$\pm$ 5.33} & 36.77 \tiny{$\pm$ 10.62} & 50.21 \tiny{$\pm$ 09.63} & \textbf{52.13} \tiny{$\pm$ 10.18} \\
 & & 8 & 34.49 \tiny{$\pm$ 08.72} & 43.04 \tiny{$\pm$ 5.22} & 48.35 \tiny{$\pm$ 7.87} & 47.98 \tiny{$\pm$ 09.73} & 53.72 \tiny{$\pm$ 10.31} & \textbf{58.13} \tiny{$\pm$ 07.28} \\
 & & 16 & 47.94 \tiny{$\pm$ 08.28} & 46.82 \tiny{$\pm$ 3.94} & 52.94 \tiny{$\pm$ 7.14} & 53.79 \tiny{$\pm$ 09.47} & 57.00 \tiny{$\pm$ 08.69} & \textbf{61.02} \tiny{$\pm$ 05.55} \\
 & & 32 & 50.80 \tiny{$\pm$ 04.52} & 51.71 \tiny{$\pm$ 4.64} & 54.26 \tiny{$\pm$ 6.37} & 53.23 \tiny{$\pm$ 5.14} & 61.12 \tiny{$\pm$ 04.83} & \textbf{64.69} \tiny{$\pm$ 02.40} \\
\Xhline{3\arrayrulewidth}\rule{0mm}{3mm}
\multirow{4}{*}{Overall Average} &  & 4 & 38.13 & 40.95 & 40.13 & 40.10 & 45.99 & \textbf{48.71} \\
     & & 8 & 36.99 & 46.37 & 45.89 & 44.25 & 50.86 & \textbf{53.70}\\
 & & 16 & 48.55 & 51.61 & 49.93 & 49.07 & 55.50 & \textbf{58.41} \\
 & & 32 & 55.30 & 56.23 & 52.65 & 55.42 & 57.02 & \textbf{60.81} \\
\bottomrule
\end{tabular}
\caption{$k$-shot accuracy on novel tasks not seen in training. Models on left of separator don't use supervised data.}
\label{tab:general}
\end{table*}

Results are presented in Table \ref{tab:general}. Results on 2 domains of Rating are in Supplementary due to space limitation.
First, comparing models which don't use any supervised data, we see that on average across the 12 tasks, the meta-trained \selfsup performs better than BERT specially for small $k \in \{4, 8, 16\}$.
Interestingly, the \selfsup model which doesn't use any supervised data, also outperforms even MT-BERT models which use supervised data for multi-task training.
Next, comparing among all the models, we see that the \selfsupmulti model performs best on average across tasks.
For instance, on average 4-shot performance across tasks, \selfsupmulti provides a relative gain in accuracy of $21.4\%$ over the best performing MT-BERT baseline. 
Compared to LEOPARD, the \selfsupmulti yields consistent improvements for all $k\in \{4,8,16,32\}$
and demonstrates steady improvement in performance with increasing data ($k$).
We note that on some tasks, such as Disaster, \selfsup is better than \selfsupmulti.
We suspect negative transfer from multi-task training on these tasks as also evidenced by the drop in performance of MT-BERT.
These results show that SMLMT meta-training learns a better initial point that enables few-shot generalization.

\begin{table*}[ht!]
\centering \fontsize{8.0}{9.5}\selectfont \setlength{\tabcolsep}{0.5em}
\begin{tabular}{ccccccccc}
\Xhline{2\arrayrulewidth}
 Task & $k$ & $\mbox{BERT}_{\mbox{base}}$ & \textbf{\selfsup} & $\mbox{MT-BERT}_{\mbox{softmax}}$ & MT-BERT & $\mbox{MT-BERT}_{\mbox{reuse}}$ & LEOPARD & \textbf{\selfsupmulti} \\[3pt] \Xhline{2\arrayrulewidth}
\multirow{4}{*}{Scitail}    & 4 & 58.53 \tiny{$\pm$ 09.74} & 50.68 \tiny{$\pm$ 4.30} & 74.35 \tiny{$\pm$ 5.86} & 63.97 \tiny{$\pm$ 14.36} & 76.65 \tiny{$\pm$ 2.45} & 69.50 \tiny{$\pm$ 9.56} & \textbf{76.75} \tiny{$\pm$ 3.36} \\
 & 8 & 57.93 \tiny{$\pm$ 10.70} & 55.60 \tiny{$\pm$ 2.40} & \textbf{79.11} \tiny{$\pm$ 3.11} & 68.24 \tiny{$\pm$ 10.33} & 76.86 \tiny{$\pm$ 2.09} & 75.00 \tiny{$\pm$ 2.42} & \textbf{79.10} \tiny{$\pm$ 1.14} \\
 & 16 & 65.66 \tiny{$\pm$ 06.82} & 56.51 \tiny{$\pm$ 3.78} & 79.60 \tiny{$\pm$ 2.31} & 75.35 \tiny{$\pm$ 04.80} & 79.53 \tiny{$\pm$ 2.17} & 77.03 \tiny{$\pm$ 1.82} & \textbf{80.37} \tiny{$\pm$ 1.44} \\
 & 32 & 68.77 \tiny{$\pm$ 6.27} & 62.38 \tiny{$\pm$ 3.22} & \textbf{82.23} \tiny{$\pm$ 1.12} & 74.87 \tiny{$\pm$ 3.62} & 81.77 \tiny{$\pm$ 1.13} & 79.44 \tiny{$\pm$ 1.99} & \textbf{82.20} \tiny{$\pm$ 1.34} \\
\midrule
\multirow{4}{*}{\shortstack[l]{Amazon\\ Books}} & 4 & 54.81 \tiny{$\pm$ 3.75} & 55.68 \tiny{$\pm$ 2.56} & 68.69 \tiny{$\pm$ 5.21} & 64.93 \tiny{$\pm$ 8.65} & 74.79 \tiny{$\pm$ 6.91} & 82.54 \tiny{$\pm$ 1.33} & \textbf{84.70} \tiny{$\pm$ 0.42} \\
 & 8 & 53.54 \tiny{$\pm$ 5.17} & 60.23 \tiny{$\pm$ 5.28} & 74.86 \tiny{$\pm$ 2.17} & 67.38 \tiny{$\pm$ 9.78} & 78.21 \tiny{$\pm$ 3.49} & 83.03 \tiny{$\pm$ 1.28} & \textbf{84.85} \tiny{$\pm$ 0.52} \\
 & 16 & 65.56 \tiny{$\pm$ 4.12} & 62.92 \tiny{$\pm$ 4.39} & 74.88 \tiny{$\pm$ 4.34} & 69.65 \tiny{$\pm$ 8.94} & 78.87 \tiny{$\pm$ 3.32} & 83.33 \tiny{$\pm$ 0.79} & \textbf{85.13} \tiny{$\pm$ 0.66} \\
 & 32 & 73.54 \tiny{$\pm$ 3.44} & 71.49 \tiny{$\pm$ 4.74} & 77.51 \tiny{$\pm$ 1.14} & 78.91 \tiny{$\pm$ 1.66} & 82.23 \tiny{$\pm$ 1.10} & 83.55 \tiny{$\pm$ 0.74} & \textbf{85.27} \tiny{$\pm$ 0.36} \\
                    \midrule
\multirow{4}{*}{\shortstack[l]{Amazon\\DVD}} & 4 & 54.98 \tiny{$\pm$ 3.96} & 52.95 \tiny{$\pm$ 2.51} & 63.68 \tiny{$\pm$ 5.03} & 66.36 \tiny{$\pm$ 7.46} & 71.74 \tiny{$\pm$ 8.54} & 80.32 \tiny{$\pm$ 1.02} & \textbf{83.28} \tiny{$\pm$ 1.85}\\
& 8 & 55.63 \tiny{$\pm$ 4.34} & 54.28 \tiny{$\pm$ 4.20} & 67.54 \tiny{$\pm$ 4.06} & 68.37 \tiny{$\pm$ 6.51} & 75.36 \tiny{$\pm$ 4.86} & 80.85 \tiny{$\pm$ 1.23} & \textbf{83.91} \tiny{$\pm$ 1.14} \\
& 16 & 58.69 \tiny{$\pm$ 6.08} & 57.87 \tiny{$\pm$ 2.69} & 70.21 \tiny{$\pm$ 1.94} & 70.29 \tiny{$\pm$ 7.40} & 76.20 \tiny{$\pm$ 2.90} & 81.25 \tiny{$\pm$ 1.41} & \textbf{83.71} \tiny{$\pm$ 1.04} \\
& 32 & 66.21 \tiny{$\pm$ 5.41} & 65.09 \tiny{$\pm$ 4.37} & 70.19 \tiny{$\pm$ 2.08} & 73.45 \tiny{$\pm$ 4.37} & 79.17 \tiny{$\pm$ 1.71} & 81.54 \tiny{$\pm$ 1.33} & \textbf{84.15} \tiny{$\pm$ 0.94} \\
\bottomrule
\end{tabular}
\caption{$k$-shot domain transfer accuracy.}
\label{tab:nli}
\end{table*}
\subsubsection{Few-shot domain transfer}
\label{sec:newdomains}
The tasks considered here had another domain of a similar task in the GLUE training tasks.
Datasets used are (1) 4 domains of Amazon review sentiments \cite{blitzer2007biographies},
(2) Scitail, a scientific NLI dataset \cite{khot2018scitail}.
Results on 2 domains of Amazon are in Supplementary due to space limitation.
A relevant baseline here is MT-BERT$_{\mbox{reuse}}$ which reuses the softmax layer from the related training task. This is a prominent approach to transfer learning with pre-trained models.
Comparing \selfsupmulti with variants of MT-BERT, we see that \selfsupmulti performs comparable or better.
Comparing with LEOPARD, we see that \selfsupmulti generalizes better to new domains.
LEOPARD performs worse than \selfsupmulti on Scitail
even though the supervised tasks are biased towards NLI, with 5 of the 8 tasks being variants of NLI tasks.
This is due to meta-overfitting to the training domains in LEOPARD which is prevented through the regularization from SMLMT in \selfsupmulti.

\begin{figure*}[ht!]
    \centering
    \includegraphics[width=\linewidth,height=1.3in]{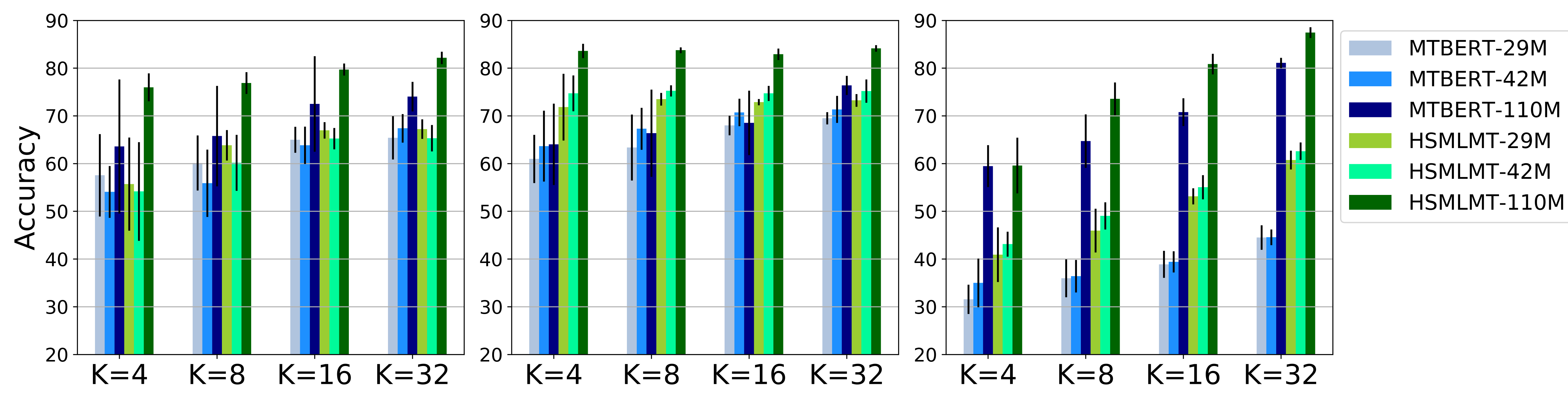}
    \caption{$k$-shot performance with number of parameters on Scitail (left), Amazon DVD (middle), and CoNLL (right). Larger models generalize better and \selfsupmulti provides accuracy gains for all parameter sizes.}
    \label{fig:params}
\end{figure*}
\begin{figure}[t!]
    \centering
    \includegraphics[width=\linewidth,trim=4 4 4 4,clip]{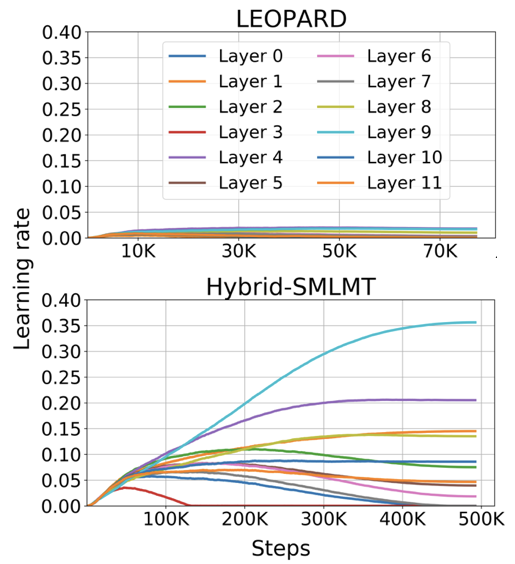}
    \caption{Learning rate trajectory during meta-training. LEOPARD learning-rates converge towards 0 for many layers, indicating meta-overfitting.}
    \label{fig:lr}
\end{figure}
\subsection{Analysis}
\textbf{Meta-overfitting:}
We study the extent of meta-overfitting in LEOPARD and \selfsupmulti.
Since these models learn the adaptation learning-rates, we can study the learning rates trajectory during meta-training.
Fig.~\ref{fig:lr} shows the results.
We expect the learning rates to converge towards zero if the task-adaptation become irrelevant due to meta-overfitting. 
LEOPARD shows clear signs of meta-overfitting with much smaller learning rates which converge towards zero for most of the layers. Note that due to this, held-out validation during training is essential to enable any generalization \cite{bansal2019learning}.
\selfsupmulti doesn't show this phenomenon for most layers and learning rates converge towards large non-zero values even when we continue training for much longer.
This indicates that SMLMT help in ameliorating meta-overfitting.

\textbf{Effect of the number of parameters:} We study how the size of the models affect few-shot performance. 
Recently, there has been increasing evidence that larger pre-trained models tend to generalize better \cite{devlin2018bert,radford2019language,raffel2019exploring}.
We explore whether this is true even in the few-shot regime.
For this analysis we use the development data for 3 tasks: Scitail, Amazon DVD sentiment classification, and CoNLL entity typing.
We consider the BERT base architecture with 110M parameters, and two smaller versions made available by \citet{Turc2019WellReadSL} consisting of about 29M and 42M parameters.
We train versions of \selfsupmulti as well as MT-BERT corresponding to the smaller models.
Results are presented in Fig.~\ref{fig:params}.
Interestingly, we see that bigger models perform much better than the smaller models even when the target task had only 4 examples per class.
Moreover, we see consistent and large performance gains from the meta-learned \selfsupmulti, even for its smaller model variants.
These results indicate that meta-training helps in data-efficient learning even with smaller models,
and enables larger models to learn more generalizable representations.

\begin{figure}[t!]
    \centering
    \begin{minipage}[b]{0.9\linewidth}
    \includegraphics[width=\linewidth]{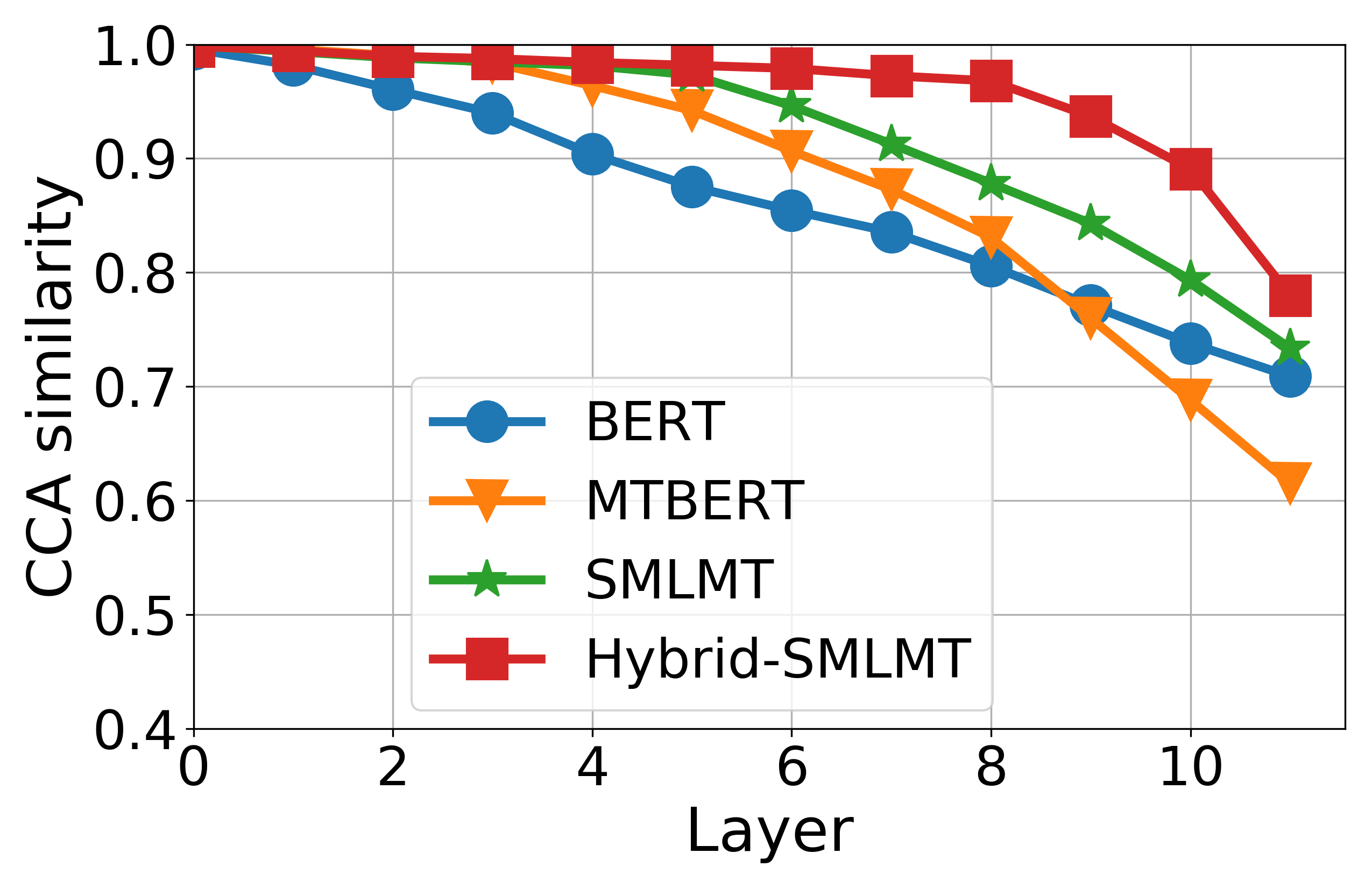}
    \end{minipage}
    \begin{minipage}[b]{0.9\linewidth}
    \includegraphics[width=\textwidth]{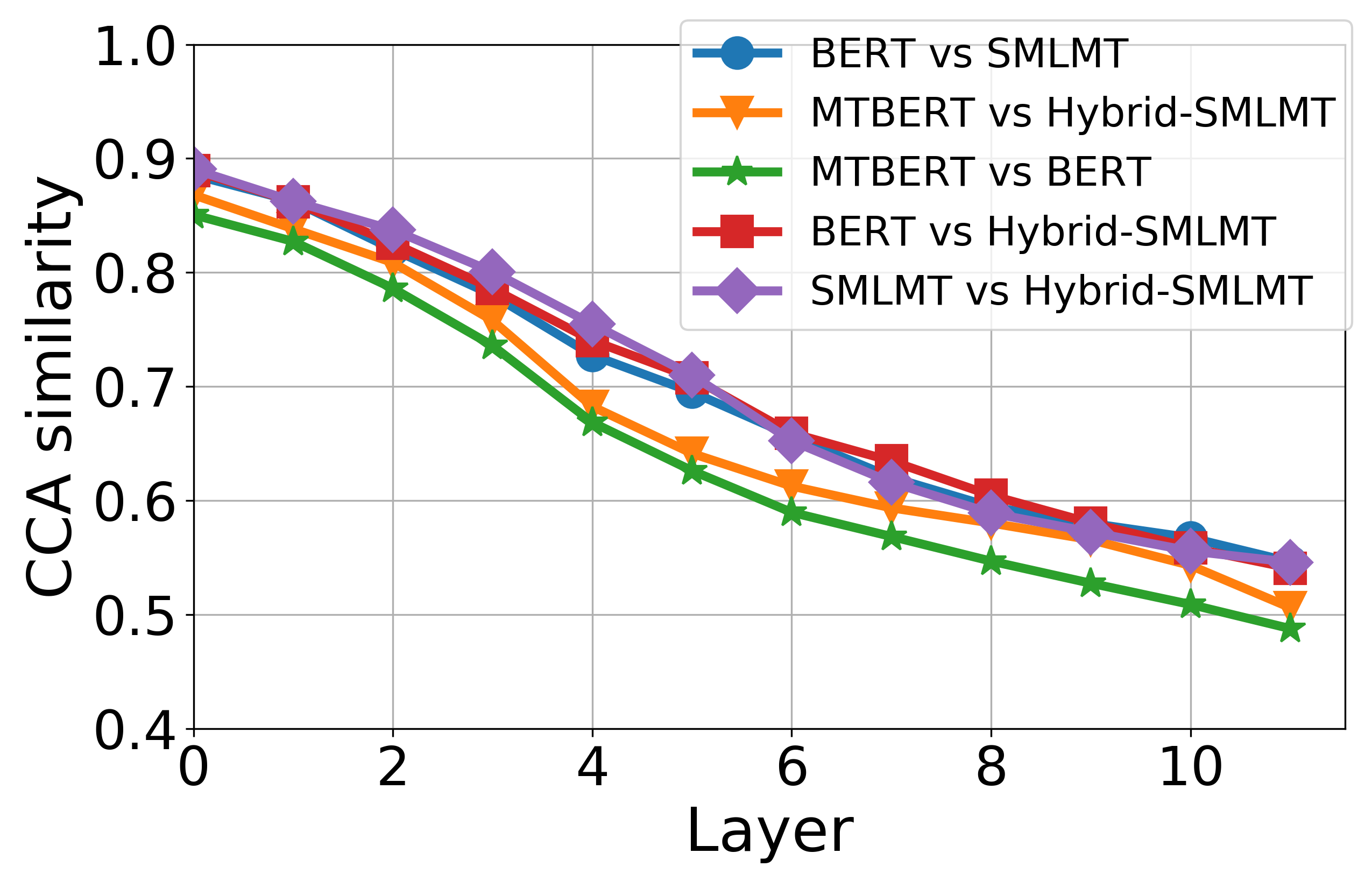}
    \end{minipage}
    \caption{CCA similarity for each transformer layer. Top: similarity before and after fine-tuning for the same model. Bottom: similarity between different pairs of models post fine-tuning. More results in Appendix.}
    \label{fig:cca}
\end{figure}
\textbf{Representation analysis:} 
To probe how the representations in the proposed models are different from the representations in the self-supervised BERT model and multi-task BERT models, we performed CCA analysis on their representations \cite{raghu2017svcca}.
We use the representations on the CoNLL and Scitail tasks for this analysis.
Results on CoNLL task are in Fig.~\ref{fig:cca}.
First, we analyze the representation of the same model before and after fine-tuning on the target task.
Interestingly, we see that the \selfsupmulti model is closer to the initial point after task-specific fine-tuning than the BERT and MT-BERT models. Coupled with the better performance of \selfsupmulti (in \ref{sec:results}), this indicates a better initialization point for \selfsupmulti.
Note that the representations in lower layers are more similar before and after fine-tuning, and lesser in the top few layers.
Next, we look at how representations differ across these models.
We see that the models converge to different representations, where the lower layer representations are more similar and they diverge as we move towards the upper layers.
In particular, note that this indicates that multi-task learning helps in learning different representations than self-supervised pre-training, and meta-learning model representations are different from the other models.

\section{Conclusion}
We introduced an approach to leverage unlabeled data to create meta-learning tasks for NLP.
This enables better representation learning, learning key hyper-parameters like learning rates, demonstrates data-efficient fine-tuning, and ameliorates meta-overfitting when combined with supervised tasks.
Through extensive experiments, we evaluated the proposed approach on few-shot generalization to novel tasks and domains and found that leveraging unlabelled data has significant benefits to enabling data-efficient generalization.
This opens up the possibility of exploring large-scale meta-learning in NLP for various meta problems, including neural architecture search, continual learning, hyper-parameter learning, and more.

\section{Acknowledgements}
This work was supported in part by the Chan Zuckerberg Initiative, in part by the National Science Foundation under Grant No. IIS-1514053 and IIS-1763618, and in part by Microsoft Research Montr\'eal. Any opinions, findings and conclusions or recommendations expressed in this material are those of the authors and do not necessarily reflect those of the sponsor.

\bibliography{references}
\bibliographystyle{acl_natbib}

\appendix
\section{Appendix}

\subsection{Training Algorithm}
The meta-training algorithm is given in \ref{alg}.
Note that $\pi^w$ are the parameters for the warp layers and $\pi$ are the remaining transformer parameters.
$L_T(\cdot)$ is the cross-entropy loss for $N$-way classification in task $T$, calculated from the following prediction:
\begin{align}
    p(y|x) &= softmax\left\{ \textbf{W}\; h_{\phi}(f_{\pi}(x)) + \textbf{b} \right\}
\end{align}
$g_{\psi}(\cdot)$ and $h_{\phi}$ are a two layer MLP with tanh non-linearity \cite{bansal2019learning}. 

\begin{algorithm}[h]
 \caption{Meta-Training}
 \begin{algorithmic}[1]

  \REQUIRE SMLMT task distribution $\mathcal{T}$ and supervised tasks $\mathcal{S}$, model parameters $\{\pi^w,\pi, \phi, \psi, \alpha\}$, adaptation steps $G$, learning-rate $\beta$, sampling ratio $\lambda$ \; \\
  Initialize $\theta$ with pre-trained BERT-base; \\
 \WHILE{not converged} 
\FOR{task\_batchsize times}
 \STATE $t \sim Bernoulli(\lambda)$
\STATE $T \sim t \cdot \mathcal{T} + (1-t) \cdot \mathcal{S}$
 \STATE $\mathcal{D}^{tr} = \{(x_j, y_j)\} \sim T$ 
 \STATE $C^n \leftarrow \{x_j|y_j=n\}$; \quad $N \leftarrow |C^n|$
 \STATE $w^n, b^n \leftarrow \frac{1}{|C^n|} \sum_{x_j \in C^n} g_{\psi}(f_{\pi}(\mathcal{D}^{tr}))$ 
 \STATE $\textbf{W} \leftarrow [w^1; \ldots; w^{N}] $; \;\;$\textbf{b} \leftarrow [b^1; \ldots; b^{N}]$
 \STATE $\theta \leftarrow \{\pi, \phi, \textbf{W}, \textbf{b}\}$; \quad $\theta^{(0)}\leftarrow \theta$
 \STATE $\Theta \leftarrow \{\pi^w, \pi, \psi, \alpha\}$
 \STATE $\mathcal{D}^{val} \sim T$ 
 \STATE $q_T \leftarrow 0$
 \FOR{$s := 0 \ldots G-1$} 
 \STATE  $\mathcal{D}_s^{tr} \sim T$ 
\STATE $\theta^{(s+1)} \leftarrow \theta^{(s)} - \alpha \nabla_{\theta}\mathcal{L}_{T}(\{\Theta, \theta^{(s)}\}, \mathcal{D}_s^{tr})$ 
\STATE $q_T \leftarrow q_T + \nabla_{\Theta}\mathcal{L}_{T}(\{\Theta, \theta^{(s+1)}\}, \mathcal{D}^{val})$
\ENDFOR
 \ENDFOR
  
 \STATE $\Theta \leftarrow \Theta - \beta \cdot \sum_T \frac{q_T}{G}$ 
 \ENDWHILE
\end{algorithmic}
\label{alg}
\end{algorithm}

\subsection{Additional Results}
\begin{figure}[t!]
    \centering
    \includegraphics[width=\linewidth]{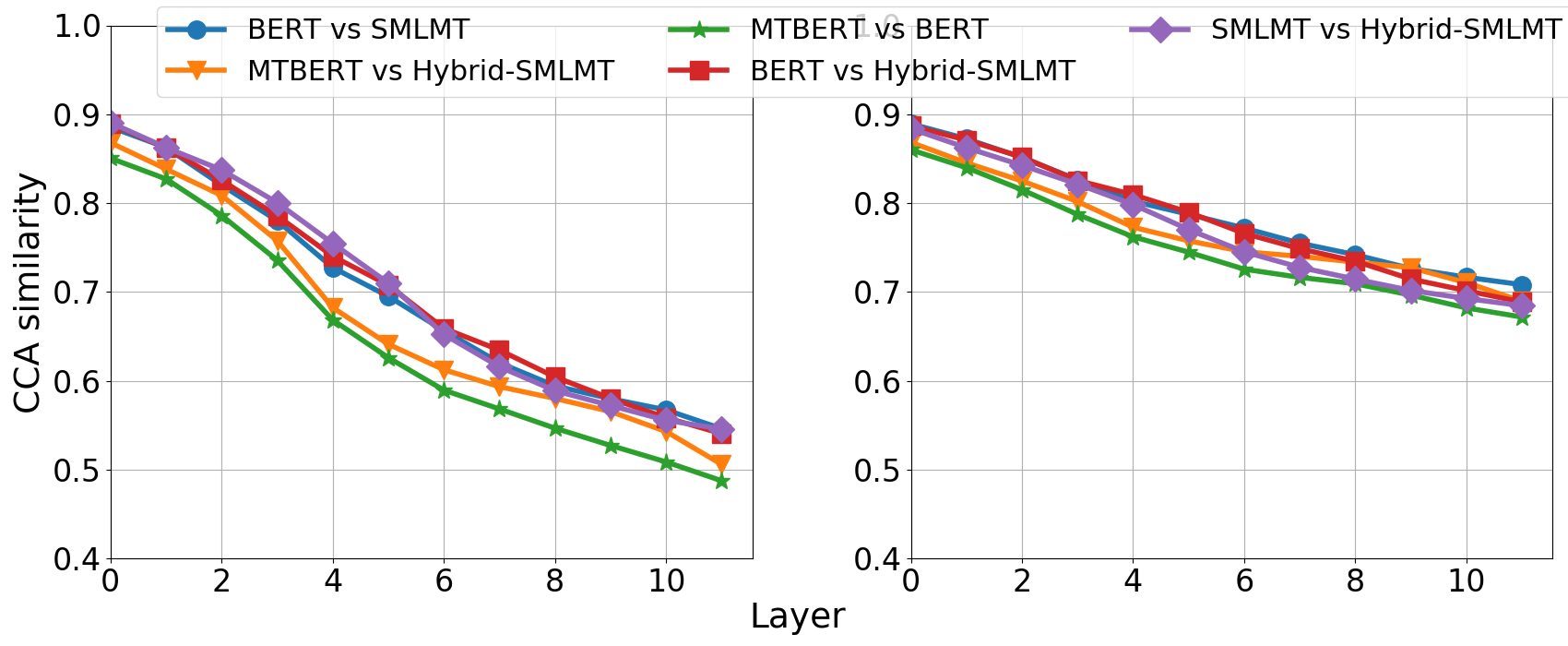}
    \caption{Cross-model CCA similarity for each layer of the transformer after fine-tuning. Left plot is on CoNLL and right on Scitail.}
    \label{fig:cca1}
\end{figure}

\begin{figure}[t!]
    \centering
    \includegraphics[width=\linewidth]{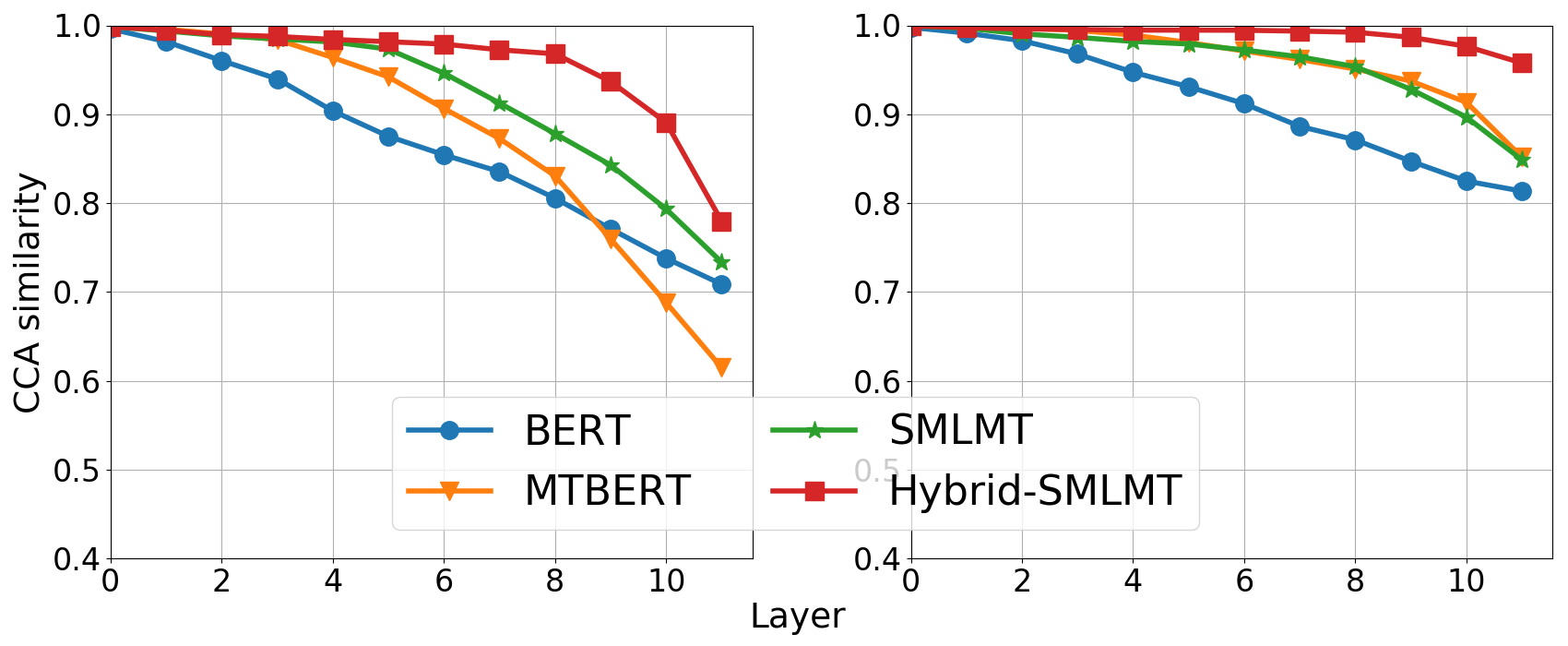}
    \caption{CCA similarity for each layer of the same model before and after fine-tuning. Left plot is on CoNLL and right on Scitail.}
    \label{fig:cca2}
\end{figure}

Table \ref{tab:supp_general} shows the results the two additional domains of Rating classification,
Table \ref{tab:supp_nli} shows the results for the two additional domains of Amazon sentiment classification.
Fig.\ref{fig:cca1} and Fig.~\ref{fig:cca2} show the CCA similarity on the two datasets: CoNLL and Scitail.
Table \ref{tab:param} shows the accuracy for different model sizes on the three evaluation datasets: Scitail, Amazon DVD, CoNLL.

\subsection{Datasets}
Dataset splits and statistics are in Table \ref{tab:dataset}.

\textbf{Supervised Training Tasks:}
We selected the GLUE \cite{wang2018glue} benchmark tasks: MRPC, SST, MNLI (m/mm), QQP, QNLI, CoLA, RTE, and SNLI \cite{bowman2015snli} as the supervised training tasks for the meta-training phase. We used the standard train/dev/test split.

\textbf{Test Tasks:}
These are same as the tasks used in \citet{bansal2019learning}. 

\subsection{Implementation Details}
\begin{table}[t]
\centering
\begin{tabular}{cc}
\Xhline{2\arrayrulewidth}
Hyper-parameter & Value \\ \Xhline{2\arrayrulewidth}
Tasks per batch & 4 \\ 
Support samples per task & 80\\ 
Query samples per task & 10\\ 
Number of classes in SSLMT & [2,3,4] \\ 
$d$ & 256  \\ 
Attention dropout & 0.1 \\ 
Hidden Layer Dropout & 0.1\\ 
Outer Loop Learning Rate & 1e-05\\ 
Adaptation Steps ($G$) & 7\\ 
$\lambda$ & 0.5\\ 
Meta-training Epochs & 1 \\
Lowercase text & False \\ 
Sequence Length & 128 \\ 
Learning-rate Warmup & 10\% of steps \\
\Xhline{2\arrayrulewidth}
\end{tabular}
\caption{Hyper-parameters.}
\label{tab:params_our}
\end{table}
\paragraph{Training Hyper-parameters:}
Table \ref{tab:params_our} lists all the hyper-parameters for the \selfsupmulti and \selfsup models. Both models use the same set of hyper-parameters, the difference being in the training tasks. Note, some hyper-parameters such as $\lambda$ are not valid for \selfsup.
We followed \citet{devlin2018bert} in setting many hyper-parameters like dropouts, and \citet{bansal2019learning} in setting hyper-parameters related to meta-learning. We use first-order MAML.
Meta-training is run for only 1 epoch, so the model always trains on a new SMLMT in every batch. This corresponds to about 500,000 steps of updates during training.

\paragraph{Sampling for \selfsupmulti :}
We restrict the word vocabulary for task creation with a term frequency of at least 50 in the corpus.
This is then used to create tasks in SMLMT as described.
This word vocabulary is discarded at this point and the data is word-piece tokenized using the BERT-base cased model vocabulary for input to the models.
Note that after a supervised task is selected to be sampled based on $\lambda$, it is sampled proportional to the square-root of the number of samples in the supervised tasks following \citet{bansal2019learning}.

\begin{table}[t!]
\centering
\resizebox{\linewidth}{!}{%
\begin{tabular}{cccccc}
\Xhline{2\arrayrulewidth}
Dataset & Labels & Train & Validation & Test \\ 
\Xhline{2\arrayrulewidth}
CoLA & 2 & 8551 & 1042 & --- \\ 
MRPC & 2 & 3669 & 409 & ---  \\ 
QNLI & 2 & 104744 & 5464 & ---  \\ 
QQP & 2 & 363847 & 40431 & ---   \\ 
RTE & 2 & 2491 & 278 & ---   \\ 
SNLI & 3 & 549368 & 9843 & ---   \\ 
SST-2 & 2 & 67350 & 873 & ---   \\ 
MNLI (m/mm) & 3 & 392703 & 19649 & ---   \\ 
Scitail & 2 & 23,596 & 1,304 & 2,126  \\ 
Amazon Sentiment Domains & 2 & 800 & 200 & 1000  \\ 
Airline & 3 & 7320 & --- & 7320   \\ 
Disaster & 2 & 4887 & --- & 4887   \\ 
Political Bias & 2 & 2500 & --- & 2500   \\ 
Political Audience & 2 & 2500 & --- & 2500 \\ 
Political Message & 9 & 2500 & --- & 2500  \\ 
Emotion & 13 & 20000 & --- & 20000  \\ 
CoNLL & 4 & 23499 & 5942 & 5648  \\ 
MIT-Restaurant & 8 & 12474 & --- & 2591   \\ 
\Xhline{2\arrayrulewidth}
\end{tabular}
}
\caption{Dataset statistics. Note that "---" indicates the correspoding split was not used. }
\label{tab:dataset}
\end{table}
\begin{table*}[htb]
\centering \fontsize{8.0}{9.5}\selectfont \setlength{\tabcolsep}{0.5em}
\begin{tabular}{ccccc|cccc}
\Xhline{2\arrayrulewidth}
    Task     & $N$ & $k$ & BERT & \textbf{\selfsup} & $\mbox{MT-BERT}_{\mbox{softmax}}$ & MT-BERT & LEOPARD & \textbf{\selfsupmulti} \\[3pt] \Xhline{2\arrayrulewidth}
\multirow{4}{*}{Rating Books} & \multirow{4}{*}{3} & 4 & 39.42 \tiny{$\pm$ 07.22} & 34.96 \tiny{$\pm$ 3.94} & 44.82 \tiny{$\pm$ 9.00} & 38.97 \tiny{$\pm$ 13.27} & 54.92 \tiny{$\pm$ 6.18} & \textbf{57.80} \tiny{$\pm$ 8.35} \\
 & & 8 & 39.55 \tiny{$\pm$ 10.01} & 37.20 \tiny{$\pm$ 4.15} & 51.14 \tiny{$\pm$ 6.78} & 46.77 \tiny{$\pm$ 14.12} & \textbf{59.16} \tiny{$\pm$ 4.13} & 56.92 \tiny{$\pm$ 5.64} \\
 & & 16 & 43.08 \tiny{$\pm$ 11.78} & 43.62 \tiny{$\pm$ 4.59} & 54.61 \tiny{$\pm$ 6.79} & 51.68 \tiny{$\pm$ 11.27} & 61.02 \tiny{$\pm$ 4.19} & \textbf{63.33} \tiny{$\pm$ 4.41} \\
 & & 32 & 52.21 \tiny{$\pm$ 4.03} & 50.45 \tiny{$\pm$ 3.28} & 54.97 \tiny{$\pm$ 6.12} & 54.95 \tiny{$\pm$ 4.82} & 64.11 \tiny{$\pm$ 2.02} & \textbf{64.51} \tiny{$\pm$ 3.06} \\
\midrule
\multirow{4}{*}{Rating DVD} & \multirow{4}{*}{3} & 4 & 32.22 \tiny{$\pm$ 08.72} & 38.26 \tiny{$\pm$ 3.62} & 45.94 \tiny{$\pm$ 7.48} & 41.23 \tiny{$\pm$ 10.98} & 49.76 \tiny{$\pm$ 9.80} & \textbf{52.08} \tiny{$\pm$ 11.03} \\
 & & 8 & 36.35 \tiny{$\pm$ 12.50} & 37.92 \tiny{$\pm$ 3.61} & 46.23 \tiny{$\pm$ 6.03} & 45.24 \tiny{$\pm$ 9.76} & \textbf{53.28} \tiny{$\pm$ 4.66} & 52.98 \tiny{$\pm$ 07.84} \\
 & & 16 & 42.79 \tiny{$\pm$ 10.18} & 41.87 \tiny{$\pm$ 4.30} & 49.23 \tiny{$\pm$ 6.68} & 45.19 \tiny{$\pm$ 11.56} & 53.52 \tiny{$\pm$ 4.77} & \textbf{56.70} \tiny{$\pm$ 04.32} \\
 & & 32 & 48.61 \tiny{$\pm$ 3.24} & 46.37 \tiny{$\pm$ 4.91} & 51.16 \tiny{$\pm$ 4.30} & 52.82 \tiny{$\pm$ 3.41} & 55.49 \tiny{$\pm$ 4.50} & \textbf{57.90} \tiny{$\pm$ 03.93} \\
\bottomrule
\end{tabular}
\caption{$k$-shot accuracy on novel tasks not seen in training. Results for 2 more rating tasks.}
\label{tab:supp_general}
\end{table*}

\begin{table*}[htb]
\centering \fontsize{8.0}{9.5}\selectfont \setlength{\tabcolsep}{0.5em}
\begin{tabular}{ccccccccc}
\Xhline{2\arrayrulewidth}
 Task & $k$ & $\mbox{BERT}_{\mbox{base}}$ & \textbf{\selfsup} & $\mbox{MT-BERT}_{\mbox{softmax}}$ & MT-BERT & $\mbox{MT-BERT}_{\mbox{reuse}}$ & LEOPARD & \textbf{\selfsupmulti} \\[3pt] \Xhline{2\arrayrulewidth}
\multirow{4}{*}{\shortstack[l]{Amazon\\ Books}} & 4 & 54.81 \tiny{$\pm$ 3.75} & 55.68 \tiny{$\pm$ 2.56} & 68.69 \tiny{$\pm$ 5.21} & 64.93 \tiny{$\pm$ 8.65} & 74.79 \tiny{$\pm$ 6.91} & 82.54 \tiny{$\pm$ 1.33} & \textbf{84.70} \tiny{$\pm$ 0.42} \\
 & 8 & 53.54 \tiny{$\pm$ 5.17} & 60.23 \tiny{$\pm$ 5.28} & 74.86 \tiny{$\pm$ 2.17} & 67.38 \tiny{$\pm$ 9.78} & 78.21 \tiny{$\pm$ 3.49} & 83.03 \tiny{$\pm$ 1.28} & \textbf{84.85} \tiny{$\pm$ 0.52} \\
 & 16 & 65.56 \tiny{$\pm$ 4.12} & 62.92 \tiny{$\pm$ 4.39} & 74.88 \tiny{$\pm$ 4.34} & 69.65 \tiny{$\pm$ 8.94} & 78.87 \tiny{$\pm$ 3.32} & 83.33 \tiny{$\pm$ 0.79} & \textbf{85.13} \tiny{$\pm$ 0.66} \\
 & 32 & 73.54 \tiny{$\pm$ 3.44} & 71.49 \tiny{$\pm$ 4.74} & 77.51 \tiny{$\pm$ 1.14} & 78.91 \tiny{$\pm$ 1.66} & 82.23 \tiny{$\pm$ 1.10} & 83.55 \tiny{$\pm$ 0.74} & \textbf{85.27} \tiny{$\pm$ 0.36} \\
                    \midrule
\multirow{4}{*}{\shortstack[l]{Amazon\\ Kitchen}} & 4 & 56.93 \tiny{$\pm$ 7.10} & 58.64 \tiny{$\pm$ 4.68} & 63.07 \tiny{$\pm$ 7.80} & 60.53 \tiny{$\pm$ 9.25} & 75.40 \tiny{$\pm$ 6.27} & 78.35 \tiny{$\pm$ 18.36} & \textbf{80.70} \tiny{$\pm$ 7.13} \\
 & 8 & 57.13 \tiny{$\pm$ 6.60} & 59.84 \tiny{$\pm$ 3.66} & 68.38 \tiny{$\pm$ 4.47} & 69.66 \tiny{$\pm$ 8.05} & 75.13 \tiny{$\pm$ 7.22} & \textbf{84.88} \tiny{$\pm$ 01.12} & 84.74 \tiny{$\pm$ 1.77} \\
 & 16 & 68.88 \tiny{$\pm$ 3.39} & 65.15 \tiny{$\pm$ 5.83} & 75.17 \tiny{$\pm$ 4.57} & 77.37 \tiny{$\pm$ 6.74} & 80.88 \tiny{$\pm$ 1.60} & 85.27 \tiny{$\pm$ 01.31} & \textbf{85.32} \tiny{$\pm$ 1.05} \\
 & 32 & 78.71 \tiny{$\pm$ 3.60} & 71.68 \tiny{$\pm$ 4.34} & 76.64 \tiny{$\pm$ 1.99} & 79.68 \tiny{$\pm$ 4.10} & 82.18 \tiny{$\pm$ 0.73} & 85.80 \tiny{$\pm$ 0.70} &  \textbf{86.33} \tiny{$\pm$ 0.67} \\
 \midrule
\multirow{4}{*}{\shortstack[l]{Amazon\\DVD}} & 4 & 54.98 \tiny{$\pm$ 3.96} & 52.95 \tiny{$\pm$ 2.51} & 63.68 \tiny{$\pm$ 5.03} & 66.36 \tiny{$\pm$ 7.46} & 71.74 \tiny{$\pm$ 8.54} & 80.32 \tiny{$\pm$ 1.02} & \textbf{83.28} \tiny{$\pm$ 1.85}\\
& 8 & 55.63 \tiny{$\pm$ 4.34} & 54.28 \tiny{$\pm$ 4.20} & 67.54 \tiny{$\pm$ 4.06} & 68.37 \tiny{$\pm$ 6.51} & 75.36 \tiny{$\pm$ 4.86} & 80.85 \tiny{$\pm$ 1.23} & \textbf{83.91} \tiny{$\pm$ 1.14} \\
& 16 & 58.69 \tiny{$\pm$ 6.08} & 57.87 \tiny{$\pm$ 2.69} & 70.21 \tiny{$\pm$ 1.94} & 70.29 \tiny{$\pm$ 7.40} & 76.20 \tiny{$\pm$ 2.90} & 81.25 \tiny{$\pm$ 1.41} & \textbf{83.71} \tiny{$\pm$ 1.04} \\
& 32 & 66.21 \tiny{$\pm$ 5.41} & 65.09 \tiny{$\pm$ 4.37} & 70.19 \tiny{$\pm$ 2.08} & 73.45 \tiny{$\pm$ 4.37} & 79.17 \tiny{$\pm$ 1.71} & 81.54 \tiny{$\pm$ 1.33} & \textbf{84.15} \tiny{$\pm$ 0.94} \\
\midrule
\multirow{4}{*}{\shortstack[l]{Amazon\\Electronics}} & 4 & 58.77 \tiny{$\pm$ 6.10} & 56.40 \tiny{$\pm$ 2.74} & 61.63 \tiny{$\pm$ 7.30} & 64.13 \tiny{$\pm$ 10.34} & 72.82 \tiny{$\pm$ 6.34} & 74.88 \tiny{$\pm$ 16.59} & \textbf{81.04} \tiny{$\pm$ 1.77}\\
 & 8 & 59.00 \tiny{$\pm$ 5.78} & 62.06 \tiny{$\pm$ 3.85} & 66.29 \tiny{$\pm$ 5.36} & 64.21 \tiny{$\pm$ 10.49} & 75.07 \tiny{$\pm$ 3.40} & 81.29 \tiny{$\pm$ 1.65} & \textbf{82.56} \tiny{$\pm$ 0.81}\\
 & 16 & 67.32 \tiny{$\pm$ 4.18} & 64.57 \tiny{$\pm$ 4.32} & 69.61 \tiny{$\pm$ 3.54} & 71.12 \tiny{$\pm$ 7.29} & 75.40 \tiny{$\pm$ 2.43} & \textbf{81.86} \tiny{$\pm$ 1.56} & 81.15 \tiny{$\pm$ 2.39}\\
 & 32 & 72.80 \tiny{$\pm$ 4.30} & 70.10 \tiny{$\pm$ 3.81} & 73.20 \tiny{$\pm$ 2.14} & 72.30 \tiny{$\pm$ 3.88} & 79.99 \tiny{$\pm$ 1.58} & 82.40 \tiny{$\pm$ 0.76} & \textbf{83.24} \tiny{$\pm$ 1.14}\\
\bottomrule
\end{tabular}
\caption{$k$-shot domain transfer accuracy for all 4 domains of Amazon sentiment classification.}
\label{tab:supp_nli}
\end{table*}
\begin{table*}[htb!]
\centering \fontsize{8.0}{9.5}\selectfont \setlength{\tabcolsep}{0.5em}
\begin{tabular}{ccccccccc}
\Xhline{2\arrayrulewidth} 
 & $k$ & \multicolumn{2}{c}{Small (29.1 M)} & \multicolumn{2}{c}{Medium (41.7 M)} & \multicolumn{2}{c}{Base (110.1 M)} & \\
 & & MT-BERT & Our & MT-BERT & Our & MT-BERT & Our \\[3pt] \Xhline{2\arrayrulewidth}
\multirow{4}{*}{Scitail}    & 4 & 57.55 \tiny{$\pm$ 8.64} & 55.70 \tiny{$\pm$ 9.75} & 54.07 \tiny{$\pm$ 5.43} & 54.17 \tiny{$\pm$ 10.34} & 63.58 \tiny{$\pm$ 14.04} &  75.98 \tiny{$\pm$ 2.93} \\
& 8 & 60.13 \tiny{$\pm$ 5.77} & 63.85 \tiny{$\pm$ 3.19} & 55.88 \tiny{$\pm$ 7.04} & 60.17 \tiny{$\pm$ 5.86} & 65.77 \tiny{$\pm$ 10.53} & 76.89 \tiny{$\pm$ 2.28} \\
& 16 & 65.00 \tiny{$\pm$ 2.73} & 66.98 \tiny{$\pm$ 1.72} & 63.84 \tiny{$\pm$ 3.91} & 65.23 \tiny{$\pm$ 2.23} & 72.50 \tiny{$\pm$ 10.01} & 79.71 \tiny{$\pm$ 1.27} \\
& 32 & 65.40 \tiny{$\pm$ 4.54} & 67.23 \tiny{$\pm$ 2.05} & 67.40 \tiny{$\pm$ 2.99} & 65.32 \tiny{$\pm$ 2.76} & 74.04 \tiny{$\pm$ 03.09} & 82.15 \tiny{$\pm$ 1.29} \\
\Xhline{2\arrayrulewidth}
\multirow{4}{*}{Amazon DVD}    & 4 & 60.99 \tiny{$\pm$ 5.05} & 71.83 \tiny{$\pm$ 6.69} & 63.66 \tiny{$\pm$ 7.43} & 74.72 \tiny{$\pm$ 3.74} & 64.04 \tiny{$\pm$ 8.53} & 83.60 \tiny{$\pm$ 1.49} \\
& 8 & 63.38 \tiny{$\pm$ 6.91} & 73.49 \tiny{$\pm$ 1.34} & 67.30 \tiny{$\pm$ 4.39} & 75.24 \tiny{$\pm$ 1.17} & 66.37 \tiny{$\pm$ 9.12} & 83.75 \tiny{$\pm$ 0.61} \\
& 16 & 67.99 \tiny{$\pm$ 2.05} & 72.88 \tiny{$\pm$ 0.66} & 70.73 \tiny{$\pm$ 2.88} & 74.72 \tiny{$\pm$ 1.58} & 68.52 \tiny{$\pm$ 6.76} & 82.91 \tiny{$\pm$ 1.20} \\
& 32 & 69.50 \tiny{$\pm$ 1.28} & 73.24 \tiny{$\pm$ 1.33} & 71.35 \tiny{$\pm$ 2.83} & 75.20 \tiny{$\pm$ 2.44} & 76.38 \tiny{$\pm$ 2.00} & 84.13 \tiny{$\pm$ 0.68} \\
\Xhline{2\arrayrulewidth}
\multirow{4}{*}{CoNLL}    & 4 & 31.57 \tiny{$\pm$ 3.06} & 40.91 \tiny{$\pm$ 5.72} & 35.00 \tiny{$\pm$ 5.11} & 43.12 \tiny{$\pm$ 2.60} & 59.47 \tiny{$\pm$ 4.40} & 59.60 \tiny{$\pm$ 5.82} \\
& 8 & 35.97 \tiny{$\pm$ 3.96} & 45.96 \tiny{$\pm$ 4.58} & 36.40 \tiny{$\pm$ 3.41} & 49.04 \tiny{$\pm$ 2.84} & 64.72 \tiny{$\pm$ 5.60} & 73.55 \tiny{$\pm$ 3.44} \\
& 16 & 38.89 \tiny{$\pm$ 2.84} & 53.14 \tiny{$\pm$ 1.70} & 39.41 \tiny{$\pm$ 2.21} & 55.05 \tiny{$\pm$ 2.54} & 70.78 \tiny{$\pm$ 2.92} & 80.85 \tiny{$\pm$ 2.15} \\
& 32 & 44.50 \tiny{$\pm$ 2.56} & 60.74 \tiny{$\pm$ 1.96} & 44.57 \tiny{$\pm$ 1.64} & 62.59 \tiny{$\pm$ 1.83} & 81.09 \tiny{$\pm$ 1.09} & 87.45 \tiny{$\pm$ 1.12} \\
\bottomrule
\end{tabular}
\caption{$k$-shot performance for three models sizes.}
\label{tab:param}
\end{table*}

\paragraph{Fine-tuning Hyper-parameter:}
We tune the number of fine-tuning epochs and batch-size using the development data of Scitail and Amazon Electronics tasks following \cite{bansal2019learning}.
Note that best values are determined for each $k$.
Epochs search range is $[5, 10, 50, 100, 150, 200, 300, 400]$ and batch-size search range is $[4, 8, 16]$.
The selected values, for $k=(4,8,16,32)$, are: (1) \selfsupmulti: epochs = $(300,350,400,200)$, batchsize = $(8,16,8,16)$; (2) \selfsup: epochs = $(100,200,150,200)$, batchsize = $(8,16,8,16)$. 
Expected overall average validation accuracy for these hyper-parameters, for $k\in (4,8,16,32)$ are: 
(1) \selfsupmulti: (0.80, 0.81, 0.83, 0.84);
(2) \selfsup: (0.54, 0.56, 0.62, 0.68).
Hyper-parameters for BERT, LEOPARD and MT-BERT are taken from \citet{bansal2019learning}.

\paragraph{Training Hardware and Time:}
We train the \selfsup and \selfsupmulti models on 4 V100 GPUs, each with 16GB memory. Owing to the warp layers, our training time per step and the GPU memory footprint is lower than LEOPARD \cite{bansal2019learning}. However, our training typically runs much longer as the model doesn't overfit unlike LEOPARD (see learning rate trajectory in main paper).
Meta-training takes a total of 11 days and 14hours.

\end{document}